%% file: main.tex
\documentclass[conference]{IEEEtran}
\let\labelindent\relax
\IEEEoverridecommandlockouts 
\usepackage{times}
\usepackage{mwe}
\usepackage[numbers]{natbib}
\usepackage{multicol}
\usepackage[bookmarks=true]{hyperref}
\usepackage{color}
\usepackage{enumitem}
\usepackage{graphicx}
\usepackage{amsmath}
\usepackage{
tikz,
relsize,
booktabs
}
\usepackage{amssymb}
\usepackage{algorithmic}
\usepackage[font={small}]{caption}
\usepackage{xcolor}
\usepackage{hyperref}
\usepackage{pifont}

\definecolor{blue}{HTML}{0173B2}
\definecolor{green}{HTML}{06A66C}
\definecolor{red}{HTML}{CF3C33}
\definecolor{magenta}{HTML}{D748A2}

\newcommand{\cmark}{{\color{green}\ding{51}}}  
\newcommand{\xmark}{{\color{red}\ding{55}}}    

\newcommand{\xxnote}[3]{}
\ifx\hidenotes\undefined
  \renewcommand{\xxnote}[3]{\color{#2}{#1: #3}}
\fi

\hypersetup{
    colorlinks=true,
    linkcolor=blue,
    filecolor=magenta,      
    citecolor=blue,
    urlcolor=teal,
}

\newcommand{\method}{\textsc{Ruka}}

\begin{document}

\title{
RUKA: Rethinking the Design of Humanoid \\Hands with Learning
}
\author{
Anya Zorin$^{\star}$
\qquad Irmak Guzey$^{\star}$
\qquad Billy Yan
\qquad Aadhithya Iyer
\\ \\ Lisa Kondrich
\qquad Nikhil X. Bhattasali
\qquad Lerrel Pinto
\\ \\ New York University
\\ {\tt \href{https://ruka-hand.github.io/}{ruka-hand.github.io}} \\
\thanks{$^{\star}$Equal contribution. Correspondence to \texttt{az61@nyu.edu}, \texttt{irmakguzey@nyu.edu}.}
}

\maketitle
\begin{abstract}
\input{0_abstract}

\end{abstract}

\IEEEpeerreviewmaketitle

\input{1_introduction}

\input{2_related_work}

\input{3_hardware_design}

\input{4_hardware_evaluation}

\input{5_controls}

\input{6_applications}

\input{7_discussion}
\input{8_acknowledgements}

\bibliographystyle{plainnat}
\bibliography{ref}

\end{document}

%% file: 0_abstract.tex
Dexterous manipulation is a fundamental capability for robotic systems, yet progress has been limited by hardware trade-offs between precision, compactness, strength, and affordability. Existing control methods impose compromises on hand designs and applications. However, learning-based approaches present opportunities to rethink these trade-offs, particularly to address challenges with tendon-driven actuation and low-cost materials. This work presents \method{}, a tendon-driven humanoid hand that is compact, affordable, and capable. Made from 3D-printed parts and off-the-shelf components, \method{} has 5 fingers with 15 underactuated degrees of freedom enabling diverse human-like grasps. Its tendon-driven actuation allows powerful grasping in a compact, human-sized form factor. To address control challenges, we learn joint-to-actuator and fingertip-to-actuator models from motion-capture data collected by the MANUS glove, leveraging the hand's morphological accuracy. Extensive evaluations demonstrate \method{}'s superior reachability, durability, and strength compared to other robotic hands. Teleoperation tasks further showcase \method{}'s dexterous movements. The open-source design and assembly instructions of \method{}, code, and data are available at \href{https://ruka-hand.github.io/}{ruka-hand.github.io}

%% file: 1_introduction.tex
\section{Introduction}

Achieving dexterity similar to human hands is essential for performing daily human tasks \cite{kumar2016dext}. In recent years, significant progress has been made in robotics toward developing autonomous dexterous policies~\cite{guzey2023dexterity,guzey2023tavi,openai2019learning,nvidia2022dextreme}. Many of these advances have been achieved through learning-based approaches, including sim-to-real methods~\cite{wang2024lessons, ma2023eureka} and imitation learning methods using teleoperated robot demonstrations~\cite{pari2021surprising, iyer2024openteach, arunachalam2022holodex} or human hand demonstrations~\cite{guzey2024bridging, wang2024dexcap}. Robotic hands are now being applied in dexterous, multimodal~\cite{yuan2024robot, bhirangi2022all}, and long-horizon~\cite{wang2023hierarchical, chen2023sequential}  tasks.

While these achievements are remarkable, progress has been limited by hardware. An ideal robotic hand must balance precision, compactness, strength, and affordability. Yet these requirements remain challenging to achieve simultaneously, so existing hand designs have needed to make trade-offs based on the available control methods and target applications. To prioritize precision, some hands integrate motors with encoders directly in their joints, but this significantly increases the hand's size and weight~\cite{shaw2023leap, allegro}. To prioritize compactness and strength, other hands adopt a tendon-driven design with actuators outside the hand~\cite{shadowhand, clone}, but this introduces nonlinearities, elasticities, and uncertainties into the force transmission system, making it challenging to model how actuators affect joint angles and fingertip positions. This challenge can be addressed to an extent by incorporating small position encoders into the hand~\cite{shadowhand}, but this is expensive, prone to mechanical failure, and difficult to repair.

\begin{figure}
    \centering
    \includegraphics[width=\linewidth]{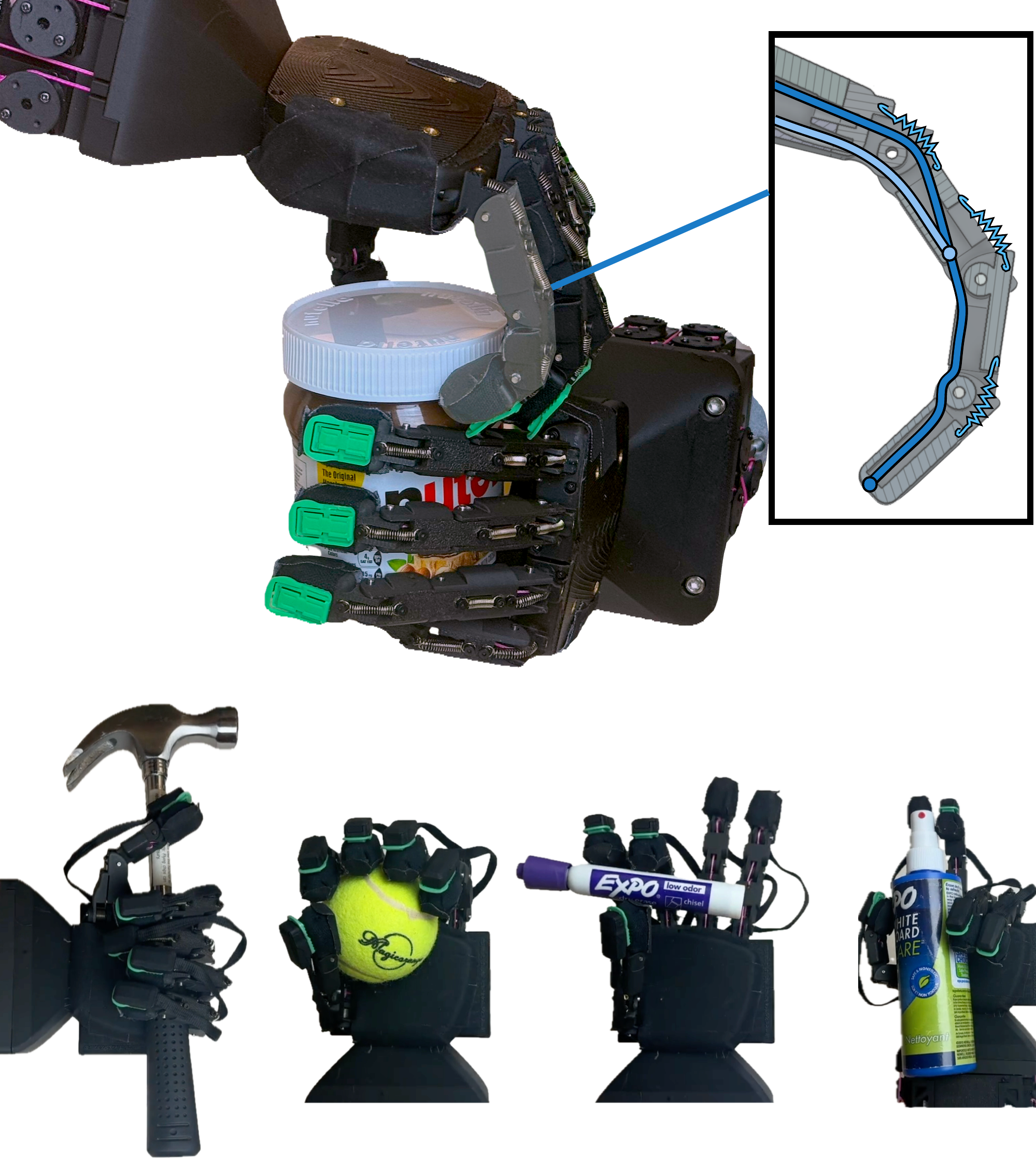}
    \caption{
        \method{} is a tendon-driven humanoid hand that is simple, affordable, and capable. Its size and morphology closely match those of a human hand, enabling it to perform diverse human-like power, precision, and fine-grained grasps.
    }
    \label{fig:intro}
\end{figure}

\begin{figure*}[t]
    \centering
    \includegraphics[width=\linewidth]{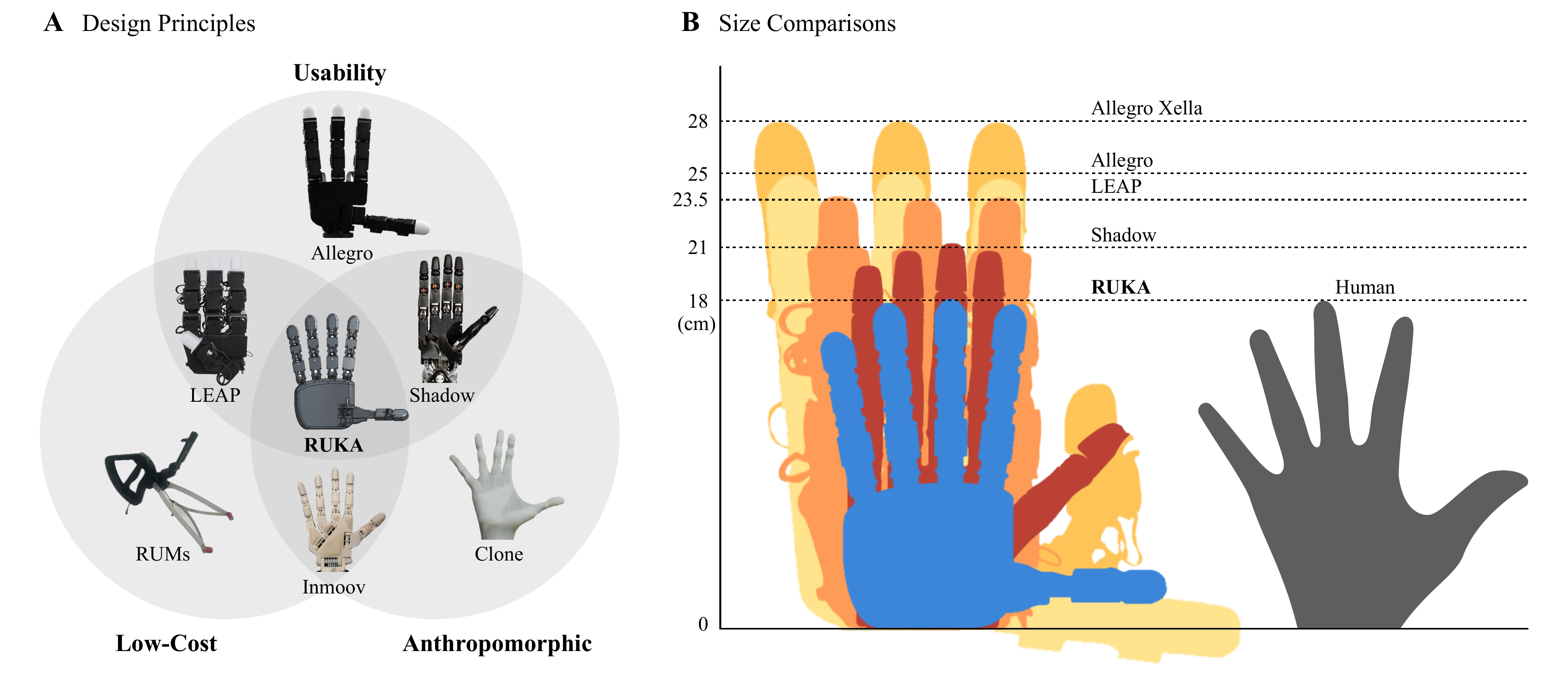}
    \caption{
        (\textbf{A}) A Venn diagram of a variety of robotic hands \cite{allegro,shaw2023leap,shadowhand,etukuru2024robot,inmoov,clone} demonstrates \method{}'s unique combination of low cost, anthropomorphism and usability.
        (\textbf{B}) An illustration of the sizes of different hands that are commonly used by the robotics community. \method{} is designed to closely match the average human hand.
    } 
    \label{fig:size_comparison}
\end{figure*}

Some degree of trade-offs is inevitable given current sensing and actuation technologies. However, we argue that learning-based approaches present an opportunity to rethink some of these trade-offs, particularly to tackle the challenges associated with tendon-driven actuation using low-cost materials. 

In this work, we introduce \method{}, a tendon-driven humanoid hand that is simple, affordable, and capable. Made from 3D-printed parts and off-the-shelf components, it takes approximately 7 hours to assemble and costs under \$1300~USD. \method{} has 5 fingers, including an opposable thumb, with dimensions matching the average human hand. These features enable smoother learning from human demonstrations and integrating into human environments. Its 15 degrees of freedom (DOFs) are driven by 11 actuators housed in the forearm and connected to finger linkages using flexible tendons. This tendon-driven actuation allows for diverse, human-like, powerful grasps in a compact, human-sized form factor. To tackle the associated control challenges, we use learning-based techniques to develop fingertip-to-actuator and joint-to-actuator models that predict the actuation commands required to produce target fingertip and joint positions. Training these models involves two key ideas: (1) We fit MANUS gloves~\cite{manus}---off-the-shelf motion-capture gloves designed for human hands---onto our robot hands to collect fingertip and joint positions without using joint encoders. This crucially leverages our hand's morphological similarity to human hands. (2) We autonomously collect labeled data points in the form of (fingertip/joint position, actuation command) pairs by procedurally sampling actuation commands within motor limits. This method captures the broad dataset needed for the models to learn these nonlinear relationships.

We extensively evaluate \method{} against popular robotic hands and demonstrate its superior reachability, durability, and strength. We further apply \method{} in teleoperation tasks and show that it can perform dexterous movements. The key contributions of this work are:

\begin{enumerate}
    \item \method{} provides an open-source design for a tendon-driven robotic hand that can be built for under \$1,300.
    \item \method{} introduces a data-driven control approach that leverages MANUS motion-capture gloves for data collection and learned controllers for fingertip positions and joint angles to support applications like teleoperation.
    \item \method{} outperforms popular robotic hands such as LEAP~\cite{shaw2023leap} and Allegro~\cite{allegro} across key metrics testing reachability, durability, and strength.
\end{enumerate}

%% file: 2_related_work.tex
\section{Related Work}
\subsection{Robotic Hands}

Direct-driven robotic hands like the LEAP \cite{shaw2023leap} and Allegro \cite{allegro} are popular in research due to their low cost and precise control using motors located directly in the joints. Despite its \$15,000 price, the Allegro hand has only four fingers, overheats easily, and is hard to repair. The LEAP hand improves endurance and repairability but, like other direct-drive designs, remains oversized and not human-like.

Tendon-driven hands with external actuators address these issues, but commercial options like the \$100,000 Shadow Hand \cite{shadowhand} are costly and require complex repairs. Open-source alternatives, such as the Inmoov \cite{inmoov} and Dexhand \cite{dexhand}, allow easier repairs but often lack infrastructure and precise control.

Soft robotic hands offer many advantages, including compliance and more natural grasps. The RBO3 hand is a soft, pneumatically actuated design that, while beneficial, is less precise and difficult to simulate due to its flexibility \cite{Puhlmann2022RBO3}. The ADAPT hand addresses this with a rigid link structure, compliant joints, and a compliant skin \cite{junge2024robustanthropomorphicroboticmanipulation}. Though both hands are robust, neither is open-source or available to buy.

Biomimetic hand designs, which mimic the exact tendon actuation of a human hand, are often extremely complex but robust. Hands such as the FLLEX hand \cite{Kim2019fluidlubricateddexterousfingermechanism} and the hand in Xu et al. \cite{Todorov2016highlybiomimetichand} use complex pulley systems and custom parts. They are highly precise while remaining compliant. However, because of their complex nature, they are difficult to reproduce. In contrast, \method{} is a compact, tendon-driven, anthropomorphic hand with a simple, accessible design, making it an effective research tool.

\begin{table}[h]
\caption{Comparison of \method{} with LEAP, Allegro, Allegro Xella, Inmoov, Shadow robotic hands \cite{shaw2023leap, allegro, shadowhand} and a human hand baseline. Evaluation across cost, degrees of freedom (DOF), degrees of actuation (DOA), actuation type (direct-drive or tendon-driven), and whether the design is open-source.}
    \begin{tabular}{l|rcccc}
    \toprule
    \textbf{Robot Hand} & \textbf{Cost}  & \textbf{DOF} & \textbf{DOA} & \textbf{Actuation} & \textbf{Open-Source} \\
    \midrule
    Human               & $-$              & 22           & $-$            & Tendon             & $-$                    \\
    LEAP                & \$2,000         & 16           & 16           &   Direct                 &  \cmark                     \\
    Allegro             & \$15,000       & 16           & 16           &      Direct              &  \xmark                   \\
    Allegro  Xella           & \$75,000       & 16           & 16           &      Direct              &  \xmark                  \\
    Inmoov              & \$100          &  14          &5             &  Tendon            &    \cmark                  \\
    Shadow              & \$100,000      &   22         &  20          &  Tendon                  &    \xmark                 \\
    \textbf{\method{}}       & \textbf{\$1,300} & \textbf{15}    & \textbf{11}    & \textbf{Tendon}          & \textbf{\cmark}            \\
    \bottomrule
    \end{tabular}
     
    \label{tab:general_comparison}
\end{table}

\subsection{Controllers for Hands}
Traditionally, hand controllers use well-defined kinematic models to control joint angles or end-effector positions. The LEAP \cite{shaw2023leap} and Allegro \cite{allegro} hands employ simple controllers where motor positions directly dictate joint angles. The HRI hand \cite{PARK2020HRIHand} uses a single motor to drive a rigid two-four-bar linkage, allowing geometric calculation of fingertip positions. The Shadow Hand \cite{shadowhand} features joint encoders for closed-loop control, while the Faive Hand \cite{Toshimitsu2023gettingtheballrolling} estimates joint angles based on tendon displacement.

When kinematics and proprioception are ill-defined, as in soft and tendon-driven hands, prior work has turned to data-driven approaches for learning controllers. While these methods have achieved great success in policy learning for dexterous manipulation~\cite{guzey2023dexterity, guzey2023tavi, guzey2024bridging, haldar2023teachrobotfishversatile, chen2023sequentialdexteritychainingdexterous, openai2019learning, chen2024object}, applying them to controller learning remains challenging due to difficulties in collecting ground-truth supervised data.

Existing approaches~\cite{Giorelli2015NNforsolvinginversestatics, Rolf2014IKonbionictrunk, Schlagenhauf2018ControlofTendon-DrivenSoftFoamRobotHands, shaw2023leap} rely on infrared sensors, Vicon motion capture, or AR tags to collect robot position data, but these methods are rigid and labor-intensive. Motion capture gloves have also been used to collect human motion data~\cite{Schlagenhauf2018ControlofTendon-DrivenSoftFoamRobotHands, Bauer2020Controlfoamdexmanip}, which is then retargeted to robots via AR tags or video-based pose estimation. However, these approaches are limited by their dependence on the wearer’s morphology and the need for carefully curated training poses.

Inspired by these methods, \method{} also follows a data-driven approach for learning controllers. However, unlike prior work, we enable large-scale autonomous data collection by fitting a motion-capture glove directly to the robotic hand, simplifying the process of gathering supervised data.

%% file: 3_hardware_design.tex
\section{Hardware Design}
\label{sec:hardware_design}
\subsection{Design Principles}
    The \method{} hand is designed for functionality and accessibility while balancing anthropomorphism, cost, and reliability.
\subsubsection{Morphologically Accurate}
    The \method{} hand mimics human morphology to enable tool use and direct application of human hand data. True anthropomorphism requires more than matching degrees of freedom—it demands accurate size, finger count, and overall form \cite{Sureshbabu2019FFPindex}. While hands like LEAP \cite{shaw2023leap} and Allegro \cite{allegro} match human degrees of freedom, they often fall short in form with fewer fingers and oversized designs. In contrast, \method{} replicates the human hand’s shape to minimize the need for retargeting human data, allowing for seamless interaction with tools and everyday objects.
\subsubsection{Low-Cost}
    Existing morphologically accurate robotic hands, such as the Shadow Hand \cite{shadowhand}, are often prohibitively expensive. For \method{} hand to serve as an accessible research tool, keeping costs low is a priority. To do so, we prefer 3D-printed parts and off-the-shelf components. The total cost of the raw materials needed, excluding tools (a 3D-printer and soldering iron), is under \$1,300 USD. There is also a \$500 and \$900 version of \method{} with varying Dynamixel motors.
\subsubsection{Reliability}
    For \method{} to serve as a reliable research tool, it must consistently reach commanded positions, operate for long durations without degradation, and be easily repairable. Rigid hinge joints are needed for repeatability, as compliance in actuation mechanisms can introduce uncertainty \cite{Morales2021Comparisonbetweenrigidandsoftpoly-articulatedprosthetichands}. However, to balance rigidity with functional flexibility, we added soft pads to the fingers, improving task performance while maintaining durability. Additionally, the hand’s open-source design allows for quick, in-house repairs, ensuring minimal downtime. The Dynamixel motors \cite{robotis} allow for accurate positioning, and the motor case has ventilation to prevent overheating during long runtimes. To ensure consistency across builds, we avoid construction methods that introduce variability, such as drilling and gluing.
\subsubsection{Open-Source}
     \method{} is fully open-source, with its 3D design and software freely available. Hands like Inmoov \cite{inmoov} lack editable CAD files, making design modification difficult. \method{} is designed in OnShape, allowing easy sharing and editing of CAD files. The control code and a MuJoCo model of the hand is available on GitHub. Open-source hardware is cheap, adaptable, and fosters collaboration among users and developers. However, it is also important to ensure easy assembly for inexperienced users and maintain documentation \cite{Patel2023penRobotHardware}. We provide step-by-step assembly instructions and repair guides. Assembling \method{} takes approximately 7 hours, and most repairs take under 20 minutes.

\begin{figure*}[t]
    \centering
    \includegraphics[width=\linewidth]{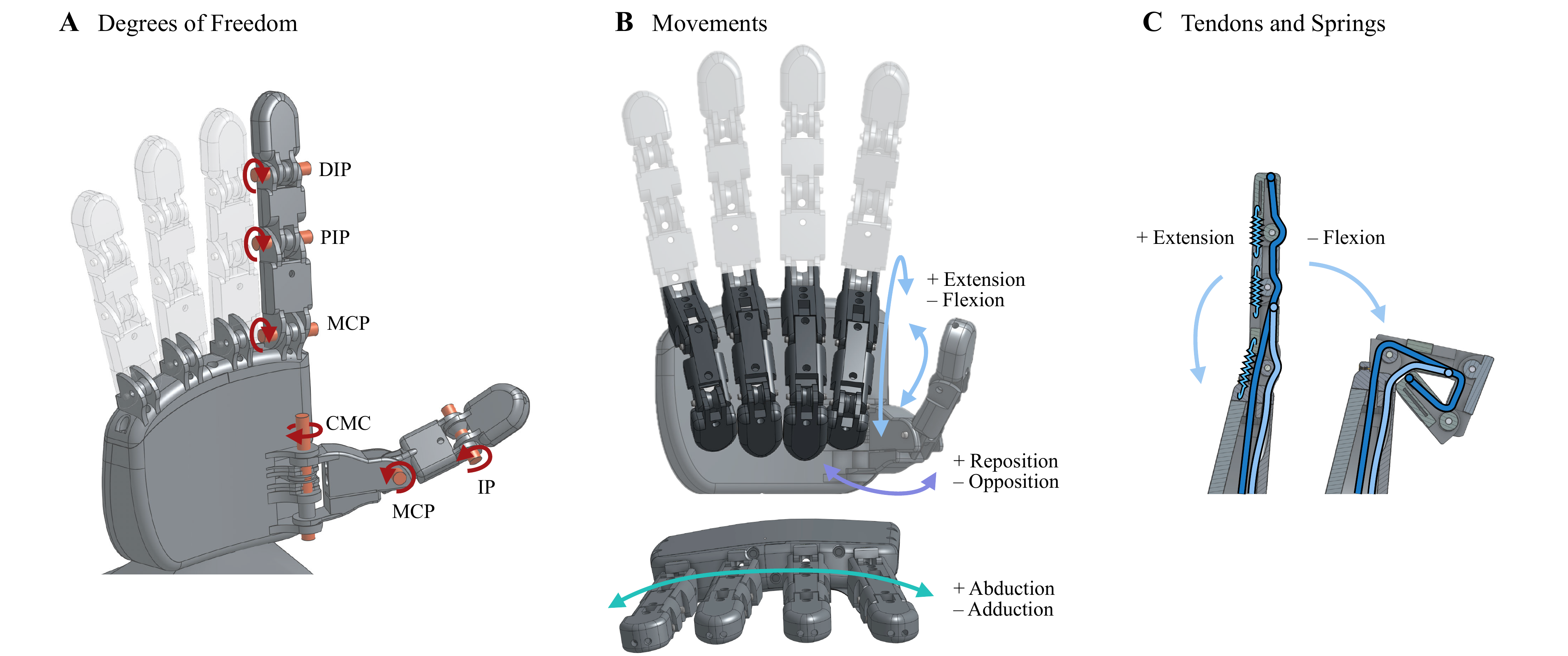}
     \caption{
     (\textbf{A}) Joints enable 15 degrees of freedom of \method{} labeled with their corresponding joint names.
     (\textbf{B}) The splay of the fingers allow for natural abduction-adduction movement without an active degree of freedom. 
     (\textbf{C}) The MCP and PIP / DIP coupled tendons (light blue and dark blue respectively) are responsible for flexion, while the springs are responsible for extension.
     }
    \label{fig:dof}
\end{figure*}
\begin{table}[b]
    \centering
    \caption{ The range of motion of each joint in \method{} in degrees compared to the human range of motion \cite{HUME1990Functionalrangeofmotionofthejointsofthehand}.}
    \begin{tabular}{lcc}
        \toprule
        \textbf{\raggedright Joint Name} & \textbf{Range of Motion} & \textbf{Human}\\ 
        \midrule
        Distal Interphalangeal (DIP) & 120\textdegree & 85\textdegree\\
        Proximal Interphalangeal (PIP) & 120\textdegree & 105\textdegree \\ 
        Metacarpophalangeal (MCP, Finger) & 140\textdegree & 85\textdegree \\ 
        \midrule
        Interphalangeal (IP) & 120\textdegree & 80\textdegree\\ 
        Metacarpophalangeal (MCP, Thumb) & 90 & 56\textdegree \\ 
        Carpometacarpal (CMC) & 190\textdegree & $-$\\  
        \bottomrule
    \end{tabular}
    
    \label{tab:range of motion}
\end{table}

\subsection{Kinematics}
The \method{} hand is designed with 15 degrees of freedom and 11 actuators. The thumb is actuated by 3 motors, 1 for each joint, while the other 4 fingers each use 2 actuators, actuating the proximal interphalangeal (PIP) and distal interphalangeal (DIP) joints together (Fig.~\ref{fig:dof}). Although the hand is underactuated, previous studies on robotic hands have demonstrated that underactuation can provide robust and reliable functionality while significantly reducing weight and mechanical complexity \cite{Jarque-Bou2019Kinematicsynergiesofhandgrasps, Matrone2010PCAmulti-dofunderactuatedhand, Puhlmann2022RBO3, Kim2019fluidlubricateddexterousfingermechanism, Toshimitsu2023gettingtheballrolling}. In human hands, DIP and PIP are rarely actuated independently \cite{Jarque-Bou2019Kinematicsynergiesofhandgrasps}, so \method{} actuates these joints with a single tendon per finger.

The metacarpophalangeal (MCP) or knuckle joint in human fingers is a ball joint, but for simplicity and rigidity, it is modeled as a revolute joint in \method{} fingers. To replicate the functions of the MCP joint, two rotations are applied to the knuckle where the joint attaches. The first rotation creates a splay, spreading the four fingers outward when the hand is fully open, allowing them to converge naturally as the hand closes, mimicing the adduction of the MCP joints (Fig.~\ref{fig:dof}). The second rotation mimics the curve of human knuckles, which originates from the carpometacarpal (CMC) joints (Fig.~\ref{fig:dof}). A concave curvature is applied to the palm and knuckles and is enhanced by slightly rotating the fingers inward toward each other. These adaptations improve the hand’s ability to achieve grasps dependent on the MCP (Fig.~\ref{fig:grasp_taxonomy}).

The thumb's kinematics are also simplified while retaining critical functionality. A human thumb typically has five degrees of freedom from three joints, enabling abduction/adduction, flexion/extension, and opposition motions. The \method{} thumb is modeled with three degrees of freedom with three joints to simplify control while optimizing grasping performance (Fig.~\ref{fig:dof}), with the ranges of motion described in Table~\ref{tab:range of motion}. The carpometacarpal (CMC) joint handles opposition. The metacarpophalangeal (MCP) joint controls abduction and adduction and is oriented 90 degrees relative to the first joint. The interphalangeal (IP) joint controls flexion and extension, rotated 45 degrees toward the palm relative to the second joint. This configuration mimics the orientation of these joints in the natural position of the thumb.

The hand is made to be similar to a human hand size. Using the average hand length and width from a dataset of hand measurements \cite{churchill1978anthropometric}, we compare \method{} to human hands and other robot hands (Fig.~\ref{fig:size_comparison}).

\subsection{Materials and Fabrication}

\subsubsection{3D-Printed Parts}
The hand's parts are 3D-printed in 24 hours using the Bambu Lab X1C \cite{bambux1c}. The finger joints, palm, wrist, and motor base are printed using PLA, chosen for its rigidity and ease of printing. The compliant pads for the fingers and palm are printed in FilaFlex Foamy TPU \cite{filaflex}. These pads are then attached using glue, and 3M friction tape is applied on top for additional grip.

\subsubsection{Off-The-Shelf Components}
Heat-set inserts are used to securely assemble the printed parts, allowing the components to be fastened with screws. These inserts are pressed into the printed parts using a soldering iron. Metal dowels are used for the pin joints, which are spring-loaded open with an extension spring. The tendons are made of braided fishing line rated for 200 lbs, selected for its combination of strength and flexibility. The line is secured to the printed components using a slip knot, routed through the finger mechanism and a PTFE tube in the palm, which guides it to the motor. The low-friction properties of PTFE significantly reduce resistance during tendon movement, ensuring smooth operation. This routing system is highly adaptable, as the only fixed points are the start and end of the PTFE tubes, allowing for easy adjustments and reconfiguration.

\subsubsection{Actuators}
The actuators are Dynamixel XM430-W210-T motors for the thumb and Dynamixel XL330-M288-T motors \cite{robotis} for the other fingers. Higher torque motors are used for the thumb. Communication with the motors is achieved through a USB-to-serial bridge. The motors are powered by a 12V and 5V dual-output power supply and connected via a custom wiring harness.

%% file: 4_hardware_evaluation.tex
\section{Hardware Evaluation}
\label{sec:hardware_evaluation}

We run a variety of tests that assess \method{} hand's capabilities and robustness. Specifically, we test its reachability, durability, and strength compared to other robot hands. 

\subsection{Reachability Tests} 

\subsubsection{Kapandji Test} We evaluate the classical Kapandji test \cite{kapandji1986}. \method{} scores 10/10 for all poses, while the Allegro and LEAP hands each score 9/10, losing a point due to having fewer fingers.

\subsubsection{Range of Motion Test} 
\begin{figure}[h]
     \centering
     \includegraphics[width=\linewidth]{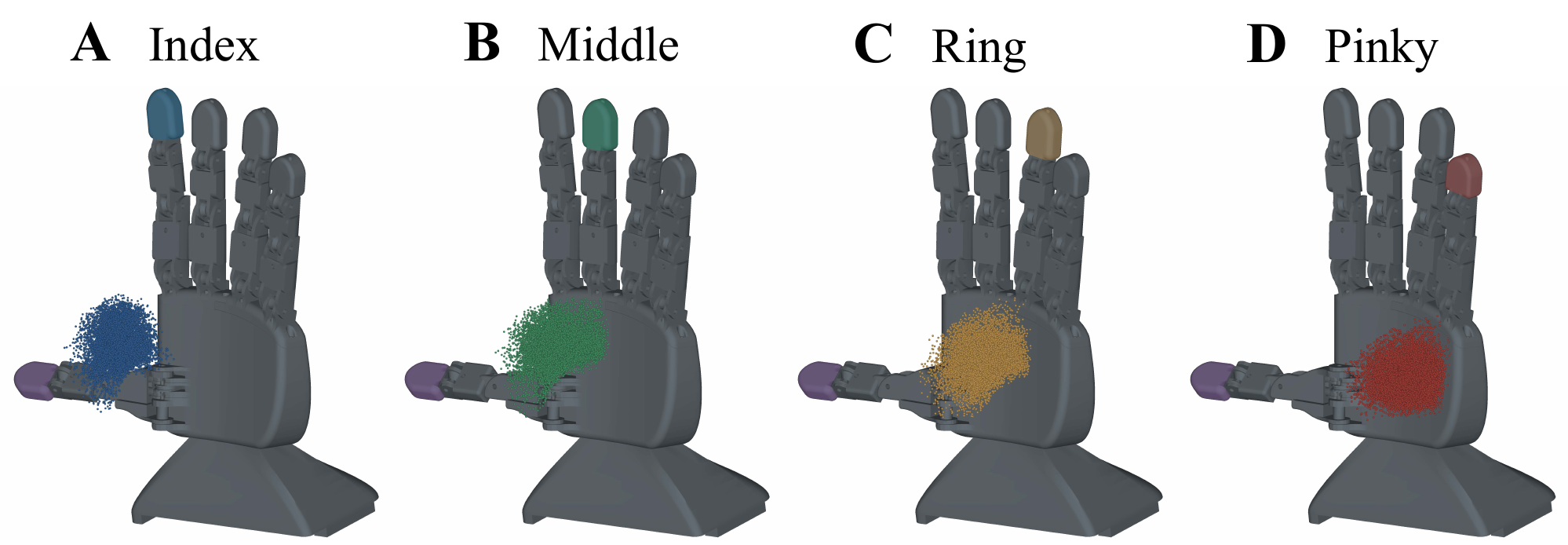}
     \caption{The intersection space of the thumb fingertip and each of the fingertips overlayed on the hand. This demonstrates the large set of opposable grasps possible with \method{}.}
     \label{fig:fingertip_space}
 \end{figure}

We evaluate the joint ranges of motion (Table~\ref{tab:range of motion}). We also evaluate the thumb's opposition capabilities by randomly sampling 250,000 joint configurations for the thumb and each finger and recording instances where the fingertips touch (Fig.~\ref{fig:fingertip_space}).

\subsubsection{Grasp Test} 
\begin{figure*}[p]
    \centering
    \includegraphics[width=\linewidth]{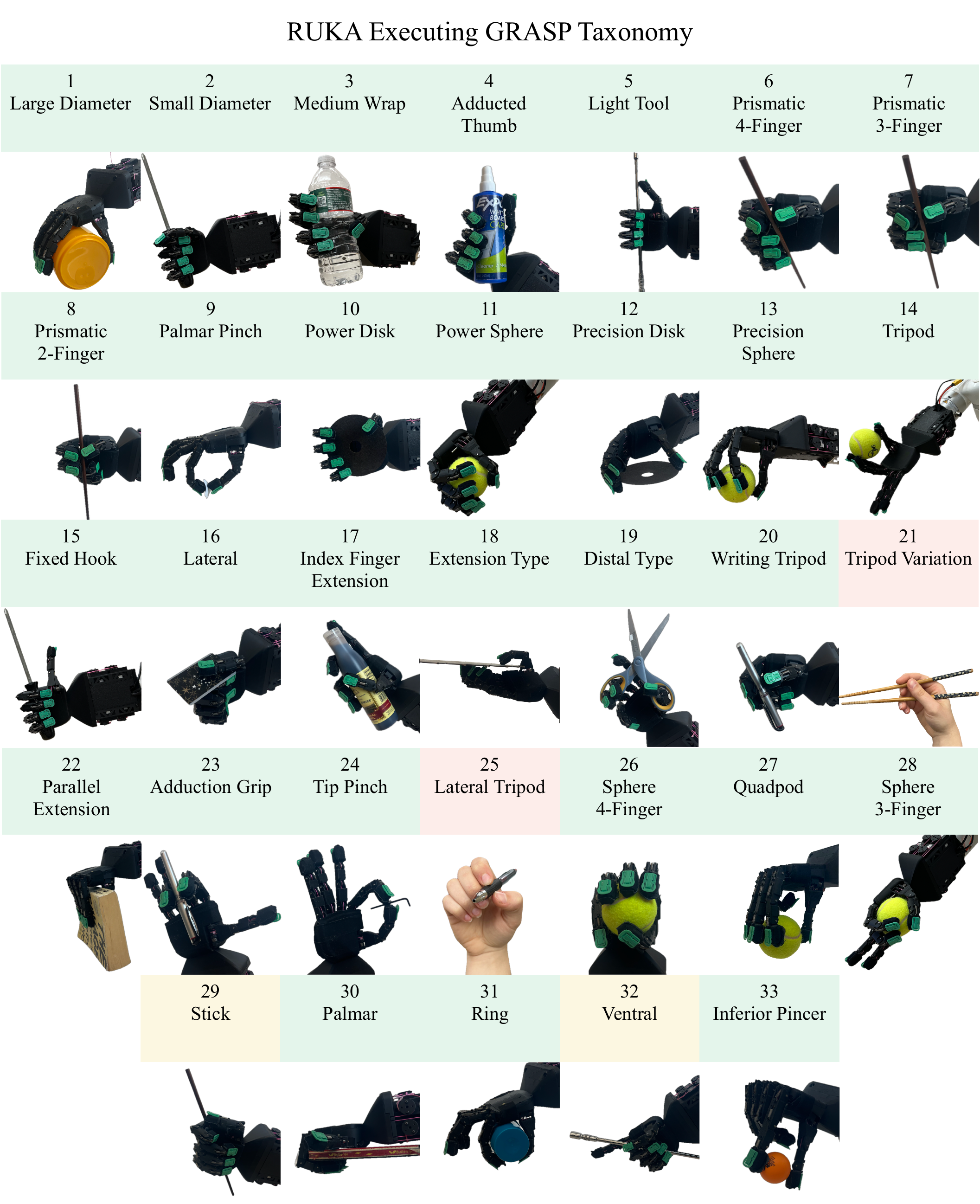}
    \caption{\method{} doing the 29 out of 33 Grasps from the GRASP Taxonomy \cite{Feix2016GRASPTaxonomy}, including a variety of power grasps, precision grasps, and intermediate grasps. Red grasps were not reached and yellow were partially reached or unstable to perturbations.} 
    \label{fig:grasp_taxonomy}
\end{figure*}

We evaluate the 33 standard grasps from the GRASP Taxonomy \cite{Feix2016GRASPTaxonomy}. \method{} successfully reproduces 29 out of 33 human hand grasps, as shown in Fig.~\ref{fig:grasp_taxonomy}. The hand is able to reach grasps that rely on MCP adduction and thumb degrees of freedom that are underactuated.

\subsection{Durability Tests}

\begin{figure}[h]
     \centering
     \includegraphics[width=\linewidth]{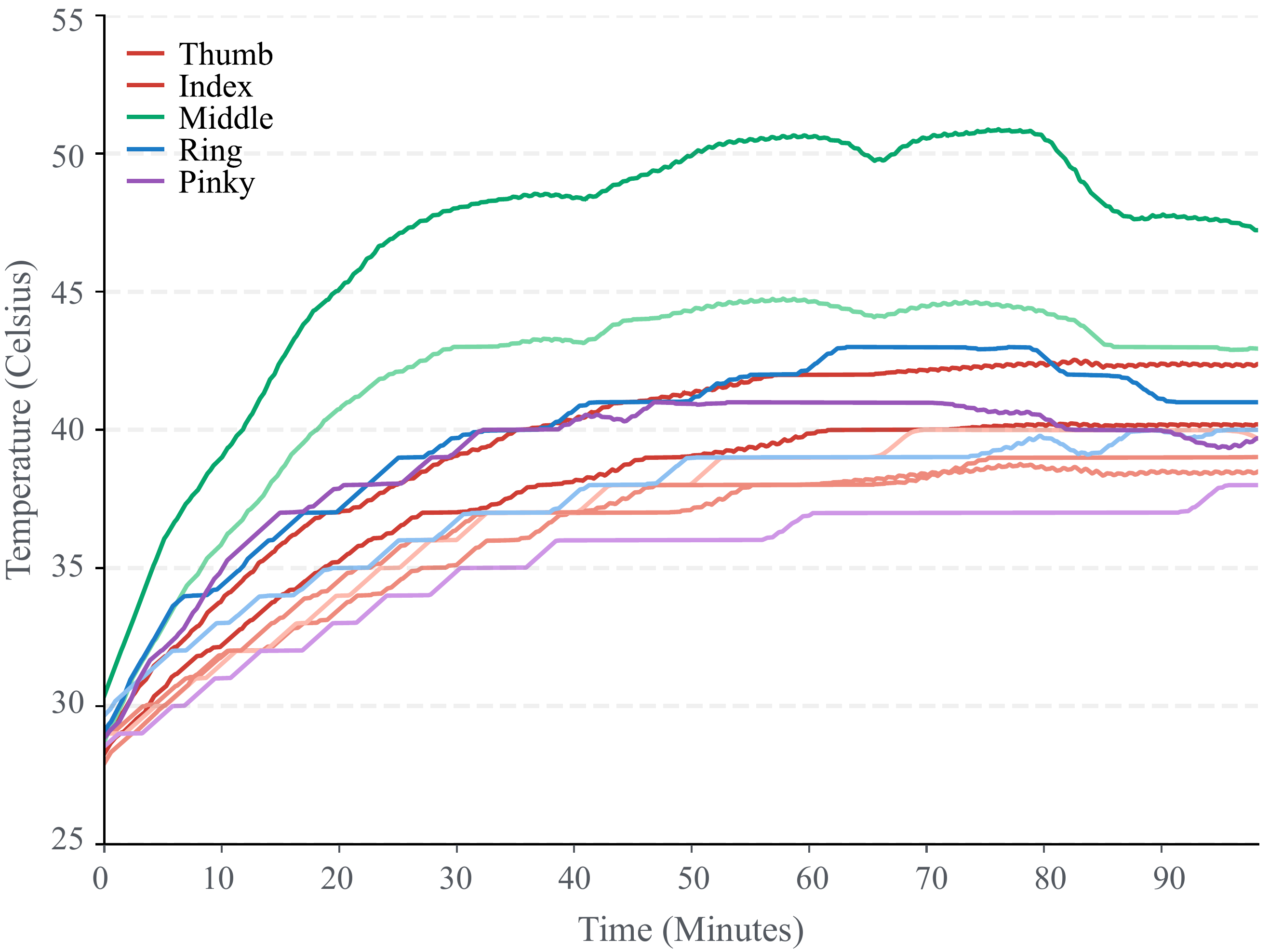}
     \caption{We run the hand continuously for 90 minutes, repeatedly doing a full range of motion. Here we show the temperature of each motor during. Note how the temperature stabilizes after some time.}
     \label{fig:temp_plot}
 \end{figure}

Durability is essential for a usable research robot. When robot hands run continuously for more than an hour, issues arise like motor overheating and declining hand precision. This means researchers cannot run continuous tasks for long periods, and the robot hand needs supervision. We were able to run the \method{} hand for 20 hours continuously, without a significant drop in motor precision. With 20 hours as a lower bound, \method{} outperforms both Allegro \cite{allegro} and LEAP \cite{shaw2023leap} hands. We also recorded the motor temperatures over 90 minutes. Even run continuously, the temperatures stabilize below the maximum rating, allowing for long run times (Fig.~\ref{fig:temp_plot}). 

\begin{figure}[h]
     \centering
     \includegraphics[width=\linewidth]{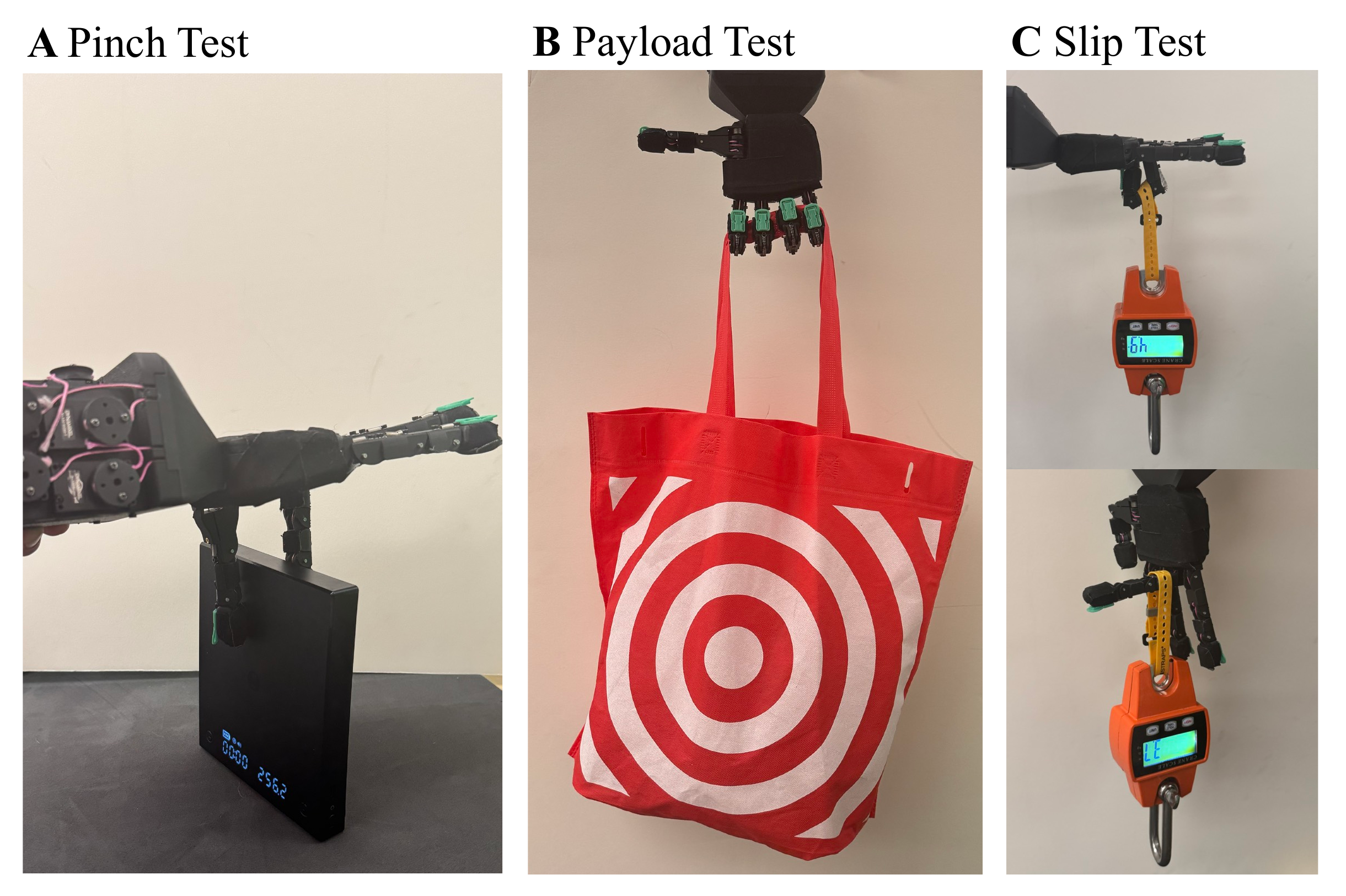}
     \caption{The experimental setup for the strength tests conducted on robot hands.}
     \label{fig:strength}
 \end{figure}

\begin{table}[h]
\caption{Results of strength tests across different robot hands.}
\centering
\begin{tabular}{@{}lcccc@{}}
\toprule
\textbf{Robot Hand} & \textbf{Pinch} (N)   & \textbf{Payload} (kg) & \textbf{DIP/PIP} (N) & \textbf{MCP} (N)\\
\midrule
Allegro Xella & 1.62  &  2.2   & 8.34   &   3.57             \\
Allegro       & 1.60  &  3.6     & 17.8  &   12.2                 \\
LEAP          & 2.45  &  4.0   & 25.17 &    11.63                  \\
Inmoov          & 2.72  &  3.2   & 15.08 & -                  \\
\textbf{\method{}} & \textbf{2.74} & \textbf{6.0} & \textbf{33.02} & \textbf{16.15} \\ 
\bottomrule
\end{tabular}
\label{tab:strength_tests}
\end{table} 
 
\subsection{Strength Tests}

\begin{figure*}[t]
    \centering
    \includegraphics[width=\linewidth]{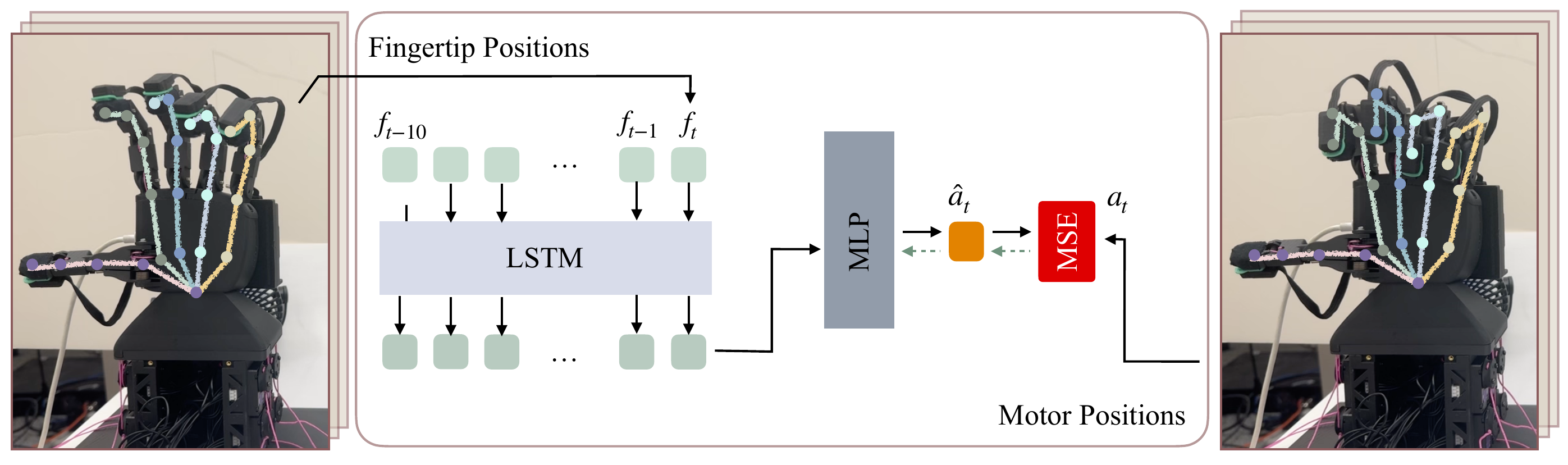}
    \caption{Keypoints received from the MANUS Haptic Gloves (left and right) and the controller architecture (center). Fingertip positions are computed from the keypoints and passed as input to an LSTM, along with the previous 10 fingertip positions. The final sequential representation from the LSTM is fed into an MLP head to predict the motor positions for each finger.} 
    \label{fig:learning}
\end{figure*}

\subsubsection{Pinch Test}
We actuate the thumb and index fingers on each hand to pinch a vertical scale measuring force. Each robot hand is tested 3 times, with the highest force recorded. The average force across left/right hands is reported in Table~\ref{tab:strength_tests}.

\subsubsection{Payload Test}
We actuate the PIP and DIP joints on the non-thumb fingers to curl around the handle of a cloth bag, to which we add weight until any joint angle error exceeds 15 degrees. The final weight is reported in Table~\ref{tab:strength_tests}. 

\subsubsection{Slip Test}
We actuate joints in combination to hold a hanging scale measuring force, to which we add weight until the joint angle error exceeds 15 degrees. We test the force supported by the combined DIP/PIP joints and the MCP joints of non-thumb fingers. The Inmoov hand \cite{inmoov} only has one actuator per finger, so DIP/PIP and MCP slip force cannot be measured; instead, the whole finger is measured jointly. The average force across the 4 fingers is reported in Table~\ref{tab:strength_tests}.
\begin{figure*}[t]
    \centering
    \includegraphics[width=\linewidth]{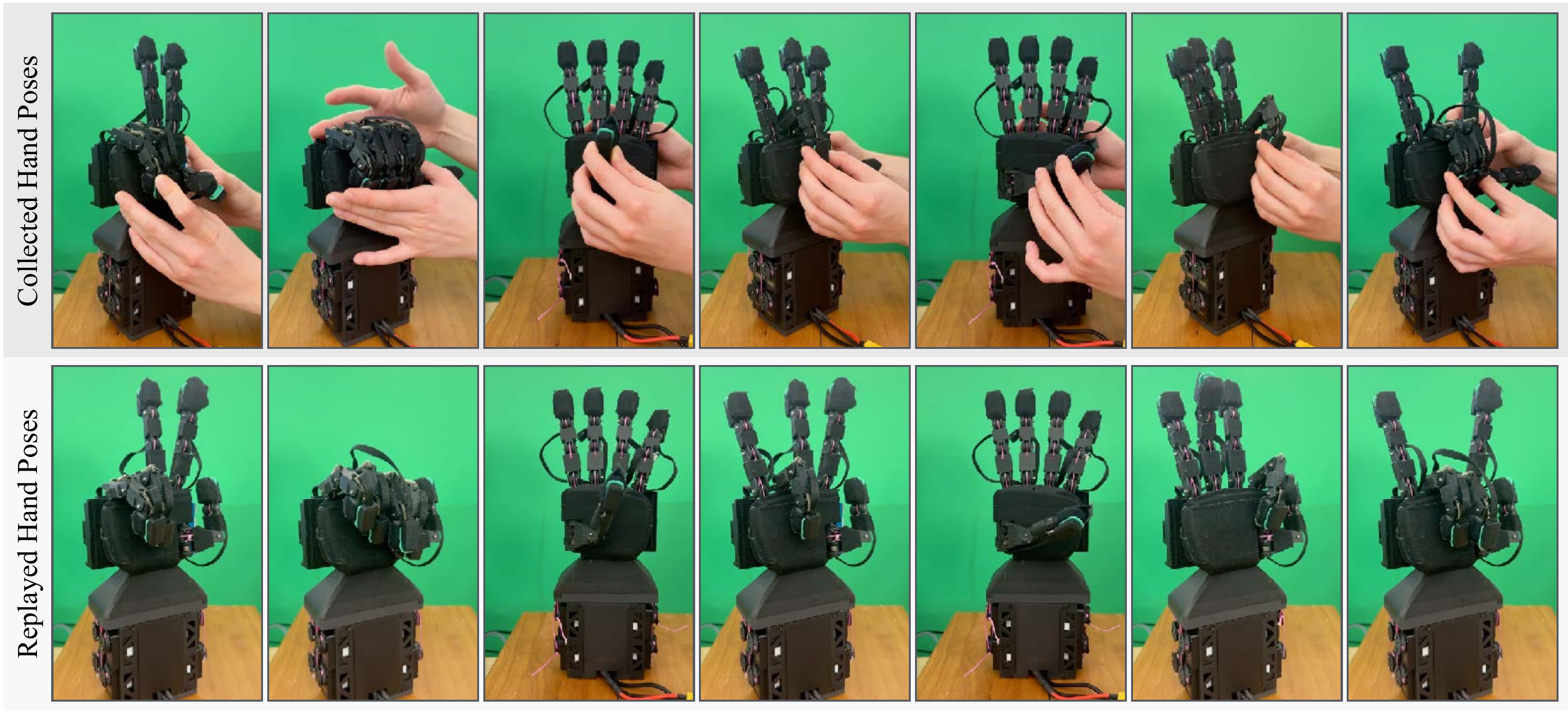}
    \caption{\method{}'s controller testing framework. The top row shows how we collect our test data on the powered-off \method{} hand and bottom row shows corresponding controller performance on the desired poses.} 
    \label{fig:testing}
\end{figure*}

\subsection{Analysis}
\method{} has the highest performance across all strength metrics. The LEAP Hand \cite{shaw2023leap} uses the same Dynamixel motors \cite{robotis} as \method{}, with three per finger instead of two for flexion. This suggests \method{}'s improved performance over LEAP is primarily due to its tendon-driven design, which removes actuator weight from the fingers. When compared to Inmoov \cite{inmoov}, which is open-source and lower cost than \method{}, \method{} outperforms Inmoov with an increased payload capacity of 87.5\% and a 64.2\%\footnote{For comparison with Inmoov, which uses only one actuator, we use \method{}'s average slip force across the DIP/PIP and MCP joints.} improvement in the slip test. This performance gain is primarily due to Inmoov’s single-actuator finger design and lower-torque motors. While Inmoov achieves a similar pinch force to RUKA, this is mainly due to Inmoov's complaint silicone fingertips, as it is only able to exert 57.4\% of force without the silicone tips. These results show that lightweight, tendon-driven designs offer significant advantages in performance.

%% file: 5_controls.tex
\section{Controller}
\label{sec:controls}

The tendon-driven approach to hand design offers significant advantages, including durability and a compact form factor, allowing robot hands to more closely mimic human hands. However, this introduces challenges in control. The absence of a direct mapping between motor positions and fingertip positions or joint angles makes it difficult to apply standard control techniques commonly used in robotics~\cite{arunachalam2022holodex, iyer2024openteach}.

In \method{}, we find the best of both worlds by employing a data-driven approach, achieving a balance between a compact, durable form and simplified control. We illustrate our framework involving the data collection process and controller learning architecture (Figure~\ref{fig:learning}). 
This section provides a detailed explanation of our data collection process and the learning methodology used for the hand controls.

\subsection{Autonomous Data Collection}
\label{sec:data_collection}

To effectively utilize a learning-based approach, we first integrate the MANUS Haptic Gloves \cite{manus} with \method{} using a custom 3D-printed attachment that secures the gloves to the back of the hand, along with the fingertip attachments provided by MANUS. The gloves track fingertip motion using magnetic field sensors paired with small embedded magnets. As the fingers move, the gloves provide real-time fingertip positions, represented as \( f_t \in \mathbb{R}^{5 \times 3} \), along with keypoint data estimating the full hand pose, \( k_t \in \mathbb{R}^{5 \times K \times 3} \), where \( K=5 \) denotes the number of keypoints per finger.

Once the gloves are attached, we initiate autonomous data collection by performing a random walk over uniformly sampled motor positions within their respective limits. For each finger, we uniformly sample from its corresponding motor positions, move the finger to that position, and then, for 100 steps, randomly increase or decrease each motor position. As described in Section~\ref{sec:hardware_design}, the thumb is actuated by three motors, whereas each of the other four fingers is actuated by two motors. To ensure coverage of the entire action space, we repeat this process 500 times for the thumb and 300 times for each of the other four fingers.

During data collection, we compute joint angles by taking the dot product between vectors connecting consecutive keypoints. Throughout data collection, we record fingertip positions, keypoints, joint angles, and both commanded and actual motor positions at 15 Hz.

\subsection{Controller Learning}

\begin{figure*}
    \centering
    \includegraphics[width=0.92\linewidth]{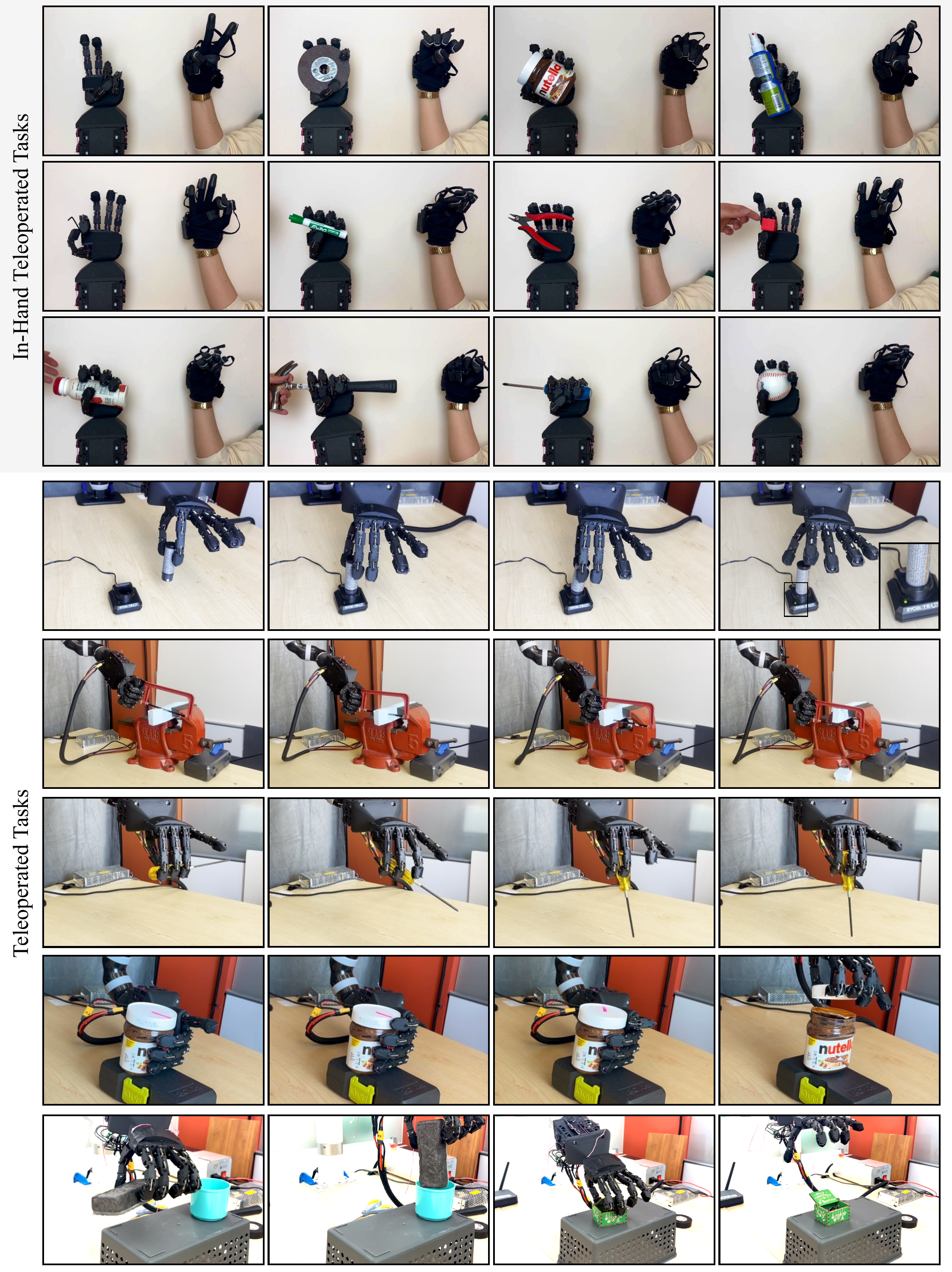}
    \caption{Teleoperated tasks made possible with \method{}. The top three rows showcase teleoperation of in-hand grasping tasks, while the remaining rows demonstrate arm-and-hand teleoperated tasks.} 
    \label{fig:teleop}
\end{figure*}

\subsubsection{Implementation Details} \label{sec:controller_implementation}
After data collection, we train a separate controller for each finger. Each controller predicts the corresponding motor positions based on the desired fingertip position for the thumb, and joint angles for the other fingers. This distinction is necessary because (a) the thumb joints differ between the human and the robot, and (b) MANUS outputs different fingertip representations for the human and the robot for the other four fingers.

We illustrate the controller learning architecture in Fig.\ref{fig:learning}. To incorporate temporal information, we use a simple Long Short-Term Memory (LSTM) \cite{lstm} network as a sequential encoder, which processes the past 10 finger states to generate a sequential representation. The final output from the LSTM is then passed to an MLP head to predict the next motor position. The loss between the predicted and ground truth motor positions is computed using mean squared error (MSE) and optimized with the AdamW~\cite{adamw} optimizer.

\subsubsection{Experiments}
We explore different architectures, input/output types, and learning parameters to optimize performance. In this section, we present our results on different data-driven controllers and the evaluations we conducted. Our experiments aim to answer the following questions: (a)~How do architecture choices and learning parameters affect controller performance? (b)~How do the trained controllers transfer to new hands?

We evaluate our approach on two separate test datasets:
\begin{itemize} 
    \item \textbf{Human Validation}: A human wears the MANUS glove, moves their hand into various poses, and the keypoints are recorded during this motion.
    \item \textbf{Robot Validation}: The MANUS glove is attached to the robot, and the robot's fingertips are manually moved while recording data from the MANUS stream, as illustrated in the top row of Fig.~\ref{fig:testing}. 
\end{itemize}

During evaluation for both datasets, we attach the glove to the robot and replay the recorded keypoints by moving the robot motors to the positions predicted by the controllers. We then record the robot’s keypoints and measure accuracy by comparing the reproduced fingertip positions against the originally saved ones.

\textbf{How do architecture choices and learning parameters affect controller performance?}

After we collected data as described in Section \ref{sec:data_collection}, we experiment on four different ways to train controllers. 

\begin{itemize}
    \item LSTM+MLP (\method{}): Controller training used in \method{} as described in Sec. \ref{sec:controller_implementation}. 
    \item MLP: Instead of integrating the past observation we directly input the current fingertip position and predict the motor position.
    \item k-Nearest Neighbor (k-NN): Instead of learning a parametric model, we use k-nearest neighbor retrieval on the desired fingertip positions and apply the mean of those neighbor's motor positions.
    \item Search-Based: Instead of training an inverse model (i.e. predicting motor positions given finger state) we train a forward model, do a search on all possible motor positions and apply the one that will give the closest predicted finger state. 
\end{itemize}

\begin{table}[ht]
\caption{Comparison of the performance of different architectures. The second column reports the mean error in Robot Validation, while the third column presents the mean error in Human Validation. Error values are computed over three positional axes for the Thumb finger.} 
\centering
\begin{tabular}{@{}lcc@{}}
\toprule
\textbf{Method} & \textbf{Error in Robot Val} (cm)   & \textbf{Error in Human Val} (cm)  \\
\midrule
Search Based  & $[\begin{array}{rrr} 0.55 & 0.59 & 0.48 \end{array}]$ & $[\begin{array}{rrr} 2.03 & 2.00 & 2.01 \end{array}]$  \\
MLP  & $[\begin{array}{rrr} 0.53 & 0.56 & 0.43 \end{array}]$ & $[\begin{array}{rrr} 1.80 & 1.90 & 2.00 \end{array}]$  \\
k-NN  & $[\begin{array}{rrr} 0.13 &  0.14 &  0.15 \end{array}]$ & $[\begin{array}{rrr} 0.95 & 0.42 & 0.83  \end{array}]$  \\
\method{} & $[\begin{array}{rrr} 0.20 & 0.27 & 0.22 \end{array}]$ & $[\begin{array}{rrr} 0.83 & 0.60 & 0.54 \end{array}]$ \\ \bottomrule
\end{tabular}
\label{tab:architecture_results}
\end{table}

Table~\ref{tab:architecture_results} presents the mean error of the thumb fingertip across different axes. We observe that while all methods perform relatively well on the Robot Validation set—reporting a maximum error of 5mm—on the Human Validation set, the MLP and search-based baselines perform significantly worse, with errors reaching up to 2cm on certain axes. Although the k-NN baseline outperforms the \method{} architecture on the Robot Validation set, it underperforms on the Human Validation set. We attribute this to the out-of-distribution (OOD) nature of human fingertip positions, which the k-NN baseline fails to generalize to effectively.

\textbf{How do the trained controllers transfer to new hands?}

Since \method{} is a tendon-driven hand that can be assembled by different users, the way its tendons are tensioned may vary from user to user. To maintain the usability of \method{} across different builds, the learned controllers should be able to generalize to these variations.

To enable this, we implement an auto-calibration script that estimates the maximum range of each motor given the current tendon tensioning. The script performs a binary search over motor positions, slowly pulling the tendons to find the point at which the finger is as curled as possible for that actuation. To ensure consistency across runs, we executed this script multiple times on the same hand and find that the variation in the maximum motor ranges found across 10 runs as 0.5 degrees.

After calibration, to evaluate how well the learned controllers generalize to a differently built hand, we repeated the same experiments from the Robot Validation set using a newly assembled and calibrated hand. The average difference in fingertip positions between the original and new hands was 3mm across different axes.
We hope that these experiments will help ensure consistency across different \method{} hands, making the system more accessible to a wider range of research labs.

%% file: 6_applications.tex
\section{Applications of \method{}}
\label{sec:applications}

\subsection{Teleoperation}
Using the controller described in Section~\ref{sec:controls}, we teleoperate \method{} to collect demonstrations for various dexterous tasks (Fig.~\ref{fig:teleop}). We use a MANUS glove for hand teleoperation and attach ArUco markers to it for arm control. By optimizing the execution time of the trained controller, ArUco detection, and \method{}’s control frequency, we achieve a teleoperation rate of 25Hz. If users prefer not to use the learned controllers and instead directly control the hand via motor positions, the system supports control at 40Hz.

Similar to the tests described in Section~\ref{sec:controls}, we input the fingertip position to the thumb controller and joint angles for the remaining fingers.

\begin{figure}[t]
    \centering
    \includegraphics[width=\linewidth]{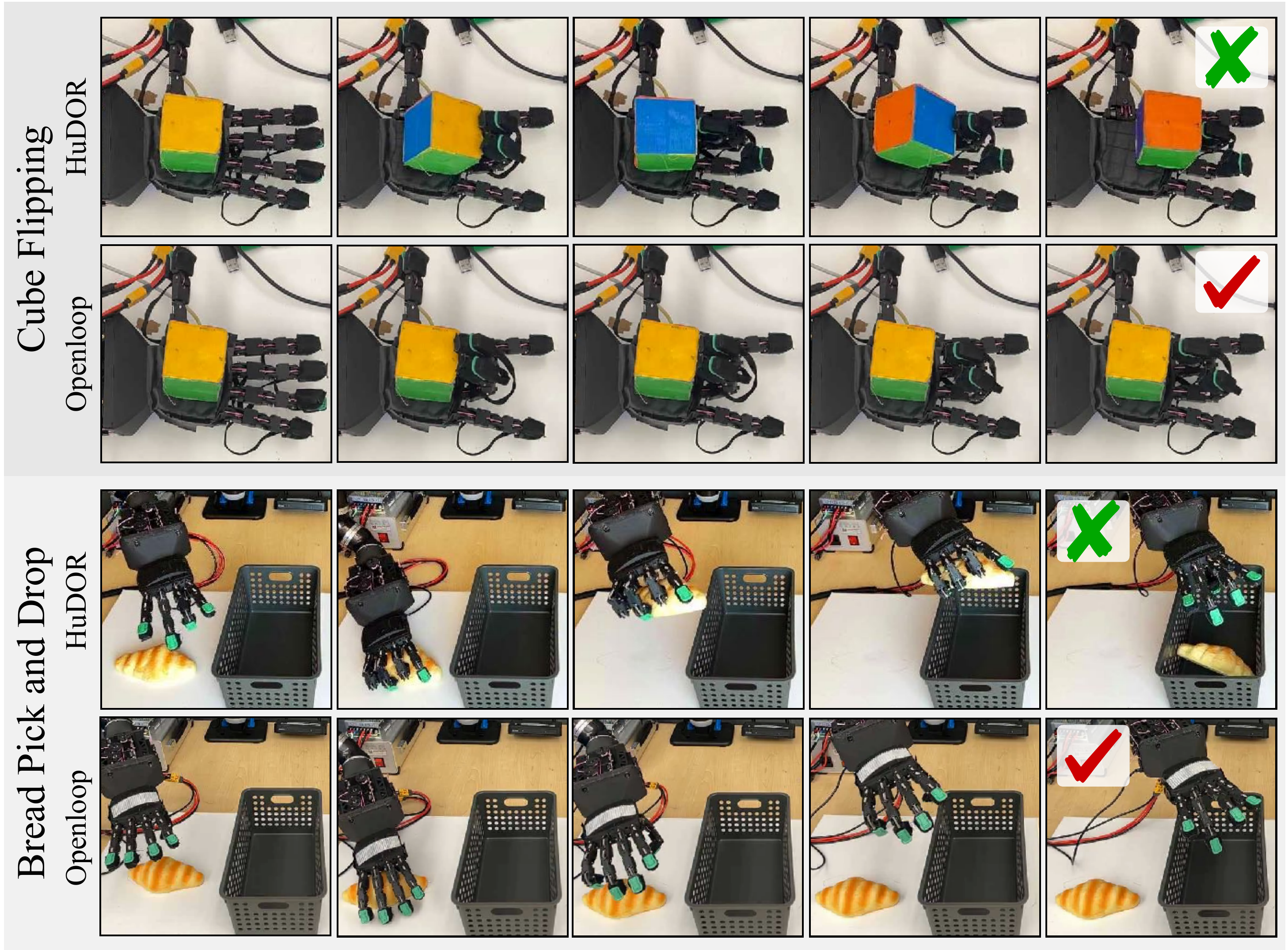}
    \caption{Rollouts of tasks learned with HuDOR~\cite{guzey2024bridging} on \method{}. For each task, the bottom row shows the open-loop rollout of the human trajectory, while the top row shows the final rollout after online finetuning, highlighting the importance of the training process.}
    \label{fig:policy_learning}
\end{figure}

\subsection{Policy Learning}

To demonstrate how \method{} can be used for policy learning approaches, we deploy HuDOR~\cite{guzey2024bridging}, an imitation learning framework that uses in-scene human videos, converts them into robot replays, and learns a residual policy to finetune the open-loop replay trajectory from the demonstrations. Rewards are computed by matching the trajectories of object centroids between the robot episodes and the human demonstrations.

We apply this approach to two tasks: \textit{Cube Flipping} and \textit{Bread Pick and Drop}. Rollouts of the policies trained using this method are shown in Fig.~\ref{fig:policy_learning}. Unlike the original HuDOR framework, for both tasks we learn a residual policy that adjusts motor positions instead of fingertip positions. On average, each policy is trained for 40 episodes, with training converging in approximately 45 minutes.

%% file: 7_discussion.tex
\section{Discussion}

In this work, we introduced \method{}, a robotic hand designed to address key challenges in existing hand designs while maintaining cost efficiency.

\textbf{Limitations:} Currently, training the model and capturing hand poses relies on the Manus glove, which provides accurate data but introduces a cost barrier for full replication and scalability. Exploring alternative motion capture techniques could improve accessibility. The current design of \method{} lacks tactile sensing, which may limit its performance in complex dexterous tasks. Integrating touch sensors could provide crucial feedback for more precise and adaptive manipulation. To maintain the current design's simplicity, it also lacks abduction at the MCP joints, which may limit the hand's ability to perform dynamic tasks. In future versions, we aim to incorporate this functionality using a compliant ball joint. 

\textbf{Hardware failure modes:} There are two primary failure modes of the hardware. First, since we use cheap Dynamixel motors with plastic gears, they wear out over time and may require replacement. The hand’s mechanism is independent of the base actuators, so this can be avoided by using metal-geared motors, albeit at a higher price point. Secondly, while the 3D-printed parts are robust to normal usage, collisions of the hand with obstacles tend to damage the MCP joint. For both failure modes, the replacement time is under 20 minutes given the simple design and easy access for hardware modifications. The design of \method{} is highly customizable, so if users require a more durable and higher-quality hand, they can adapt the design to use higher-cost components.

%% file: 8_acknowledgements.tex
\section*{Acknowledgements}
We thank Raunaq Bhirangi, Siddhant Haldar and Venkatesh Pattabiraman for valuable feedback and discussions. This work was supported by grants from Honda, Hyundai, NSF award 2339096, and ONR award N00014-22-1-2773. LP is supported by the Sloan and Packard Fellowships. NXB is supported by the Fannie and John Hertz Foundation Fellowship.